\documentclass[lettersize,journal]{IEEEtran}
\usepackage{ulem}
\usepackage{amsmath,amsfonts}
\usepackage{algorithmic}
\usepackage{algorithm}
\usepackage{array}
\usepackage{cite}
\usepackage{color}
\usepackage{multirow} 
\usepackage[caption=false,font=footnotesize,labelfont=rm,textfont=rm]{subfig}
\usepackage{textcomp}
\usepackage{stfloats}
\usepackage{url}
\usepackage{verbatim}
\usepackage{graphicx}
\usepackage{booktabs}
\usepackage{hyperref}
\usepackage{xcolor} 
\usepackage{url}

\usepackage{ulem}

\makeatletter
\newcommand{\coloredcite}[1]{{\color{blue}\cite{#1}}}
\makeatother

\hypersetup{
colorlinks=true,
linkcolor=blue,
citecolor=blue
}

\begin{document}

\title{HoloDreamer: Holistic 3D Panoramic World Generation from Text Descriptions}

\author{
Haiyang Zhou, 
Xinhua Cheng, 
Wangbo Yu, 
Yonghong Tian, and 
Li Yuan
\thanks{
Corresponding author: Yonghong Tian; Li Yuan.}
\thanks{
Haiyang Zhou is interning at Peking University,
School of Electronic and Computer Engineering, Shenzhen Graduate School, Shenzhen, Guangdong Province 518055, China (e-mail: zhouhaiyang000@gmail.com).

Xinhua Cheng is with the Peking University,
School of Electronic and Computer Engineering, Shenzhen Graduate School, Shenzhen, Guangdong Province 518055, China (e-mail: chengxinhua@stu.pku.edu.cn; yuwangbo98@gmail.com).

Wangbo Yu, Yonghong Tian, and Li Yuan are with the Peking University,
School of Electronic and Computer Engineering, Shenzhen Graduate School, Shenzhen, Guangdong Province 518055, China, and also with Peng Cheng Laboratory, Shenzhen, Guangdong Province 518066, China (e-mail: yhtian@pku.edu.cn; yuanli@u.nus.edu).
}
}

\maketitle

\begin{abstract}
3D scene generation is in high demand across various domains, including virtual reality, gaming, and the film industry. 
Owing to the powerful generative capabilities of text-to-image diffusion models that provide reliable priors, the creation of 3D scenes using only text prompts has become viable, thereby significantly advancing researches in text-driven 3D scene generation.
In order to obtain multiple-view supervision from 2D diffusion models,  prevailing methods typically employ the diffusion model to generate an initial local image, followed by iteratively outpainting the local image using diffusion models to gradually generate scenes. Nevertheless, these outpainting-based approaches prone to produce global inconsistent scene generation results without high degree of completeness, restricting their broader applications.
To tackle these problems, we introduce HoloDreamer, a framework that first generates high-definition panorama as a holistic initialization of the full 3D scene, then leverage 3D Gaussian Splatting (3D-GS) to quickly reconstruct the 3D scene, thereby facilitating the creation of view-consistent and fully enclosed 3D scenes.
Specifically, we propose Stylized Equirectangular Panorama Generation, a pipeline that combines multiple diffusion models to enable stylized and detailed equirectangular panorama generation from complex text prompts. Subsequently, Enhanced Two-Stage Panorama Reconstruction is introduced, conducting a two-stage optimization of 3D-GS to inpaint the missing region and enhance the integrity of the scene. Comprehensive experiments demonstrated that our method outperforms prior works in terms of overall visual consistency and harmony as well as reconstruction quality and rendering robustness when
generating fully enclosed scenes.
\end{abstract}


\begin{figure*}[!t]
\centering
\includegraphics[width=0.95\textwidth]{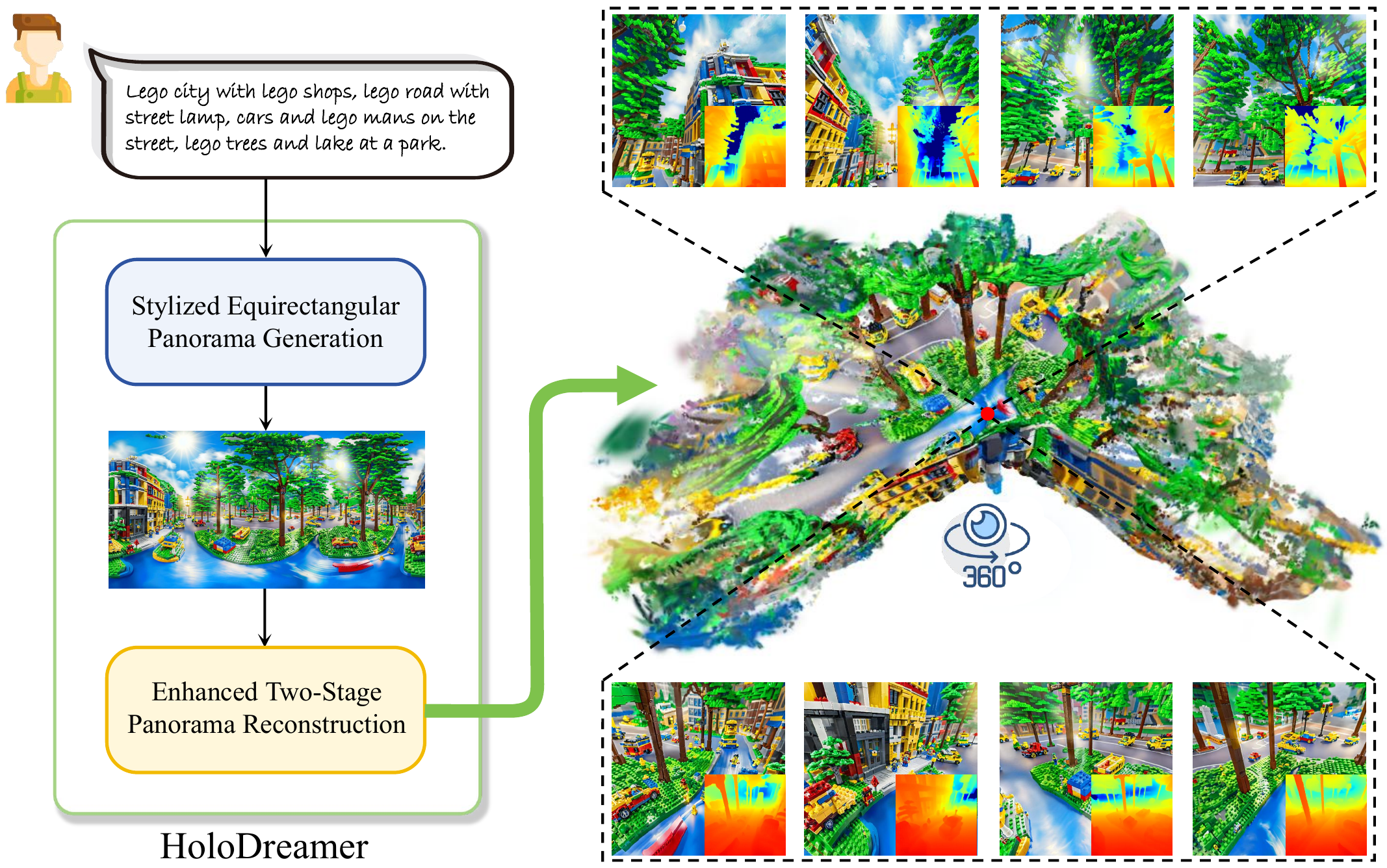}
\caption{We propose HoloDreamer, a text-driven 3D scene generation framework to generate immersive and fully enclosed 3D scenes with high view-consistency. It consists of two basic modules: Stylized Equirectangular Panorama Generation, which generates a stylized high-quality equirectangular panorama from the input user prompt, and Enhanced Two-Stage
Panorama Reconstruction, which employs 3D Gaussian Splatting for rapid 3D reconstruction of the panorama with enhanced integrity.}
\label{fig:results}
\end{figure*}

\section{Introduction}

\IEEEPARstart{A}{s} the field of 2D generation\coloredcite{rombach2021highresolution} and 3D representation evolves\coloredcite{mildenhall2020nerf, kerbl20233d}, 3D content generation has become a significant task within the realm of computer vision. Text prompts can intuitively and comprehensively describe user's needs, as a consequence, the zero-shot text-driven generation of 3D scenes will lower the barrier to entry for newcomers, and save considerable manual effort in 3D modeling. This makes it a promising application in industries such as metaverse, virtual reality and film production. However, unlike the abundance of paired text-to-image data in the field of 2D, paired text-to-3D data is significantly scarce currently. The creation of 3D datasets often requires substantial human and material resources, which results in challenges to directly train for 3D content generation from user prompts. Despite numerous efforts\coloredcite{nichol2022point,jun2023shap,liu2023meshdiffusion} to leverage diffusion models to conduct end-to-end and feedforward generation of 3D content, the results are still constrained by the size and quality of training data, leading to poor performance in details.

To overcome this limitation, some prior works\coloredcite{wang2022clip,jain2022zero,poole2023dreamfusion,wang2023score,lin2023magic3d,wang2024prolificdreamer} harness the high-level priors from pre-trained text-to-image models, \textit{i.e.}, CLIP\coloredcite{radford2021learning} and image diffusion models\coloredcite{rombach2021highresolution}, to guide the optimization of 3D representations, achieving zero-shot effects. However, these methods are limited to generating simple geometric shapes, with the cameras all converging on the object's position. For the generation of more complex scenes with camera orientations diverge outward, several past efforts, including SceneScape\coloredcite{fridman2024scenescape}, Text2Room\coloredcite{hollein2023text2room}, Text2NeRF\coloredcite{zhang2024text2nerf} and LucidDreamer\coloredcite{chung2023luciddreamer}, progressively outpaint an initial image using diffusion models, followed by the integration of monocular depth estimation networks to obtain depth information for subsequent 3D reconstruction. However, the large field of view necessitates a substantially increased number of outpainting iterations. Each iteration of the outpainting process is solely based on the local existing scene, leading to difficulties in maintaining global consistency and harmony during the prolonged outpainting process. The generated scene is visually chaotic, particularly when the scene is fully enclosed. In addition to this, the scene exhibits low rendering robustness for outside preset views.

In this work, we introduce a framework named HoloDreamer, a novel method for text-driven generation of view-consistent and fully enclosed 3D scenes with strong rendering robustness. Unlike previous approachs, which are prone to consistency issues arising from progressive outpainting, we propose Stylized Equirectangular Panorama Generation, utilizing text-to-image diffusion models to directly generate a highly consistent 360-degree equirectangular panorama from text prompts. The generated panorama boasts excellent visual quality, characterized by high-resolution details that contribute to a coherent and immersive viewing experience. 
Specifically, to preserve the accuracy of the equirectangular projection, we first generate a base panorama using a diffusion model fine-tuned on the panorama database, and subsequently perform style transfer and detail enhancement using conditional controlled diffusion models, ensuring that the panorama are not only detailed but also aesthetically pleasing and true to the visual style inferred from the text description.

We choose 3D Gaussian Splatting (3D-GS)\coloredcite{kerbl20233d} as the 3D representation for the scene due to its robust capability to represent highly granular details across various scenes and its significant optimization speed, which enables rapid, high-fidelity 3D reconstruction with a panoramic field of view. We propose Enhanced Two-Stage Panorama Reconstruction, a pipeline that reconstructs enhanced 3D scene from generated panoram using 3D-GS.
Initially, leveraging the depth prior provided by a monocular depth estimation model, we perform depth estimation on the panorama. The resulting RGBD data is then transformed into point clouds, which serve as the initialization for the 3D Gaussians.
Furthermore, to enhance the robustness of the scene rendering, a two-stage 3D-GS optimization process is designed for the reconstruction of 3D scenes. In the Pre Optimization stage, we project multiple additional perspective images from the point cloud to apply multi-view constraints on the 3D Gaussians during the optimization process. This strategy overcomes the limitation of having a single viewpoint in panorama and prevents the generation of artifacts. After Pre Optimization results, we employ a 2D inpainting model to fill in missing areas within the images rendered from the scene. Ultimately, the inpainted images are incorporated as supervision for Transfer Optimization stage to achieve high-level integrity of the final reconstructed scene.

\begin{figure*}[!t]
\centering
\subfloat[Generation Results of Indoor Scenes.]{\includegraphics[width=0.98\textwidth]{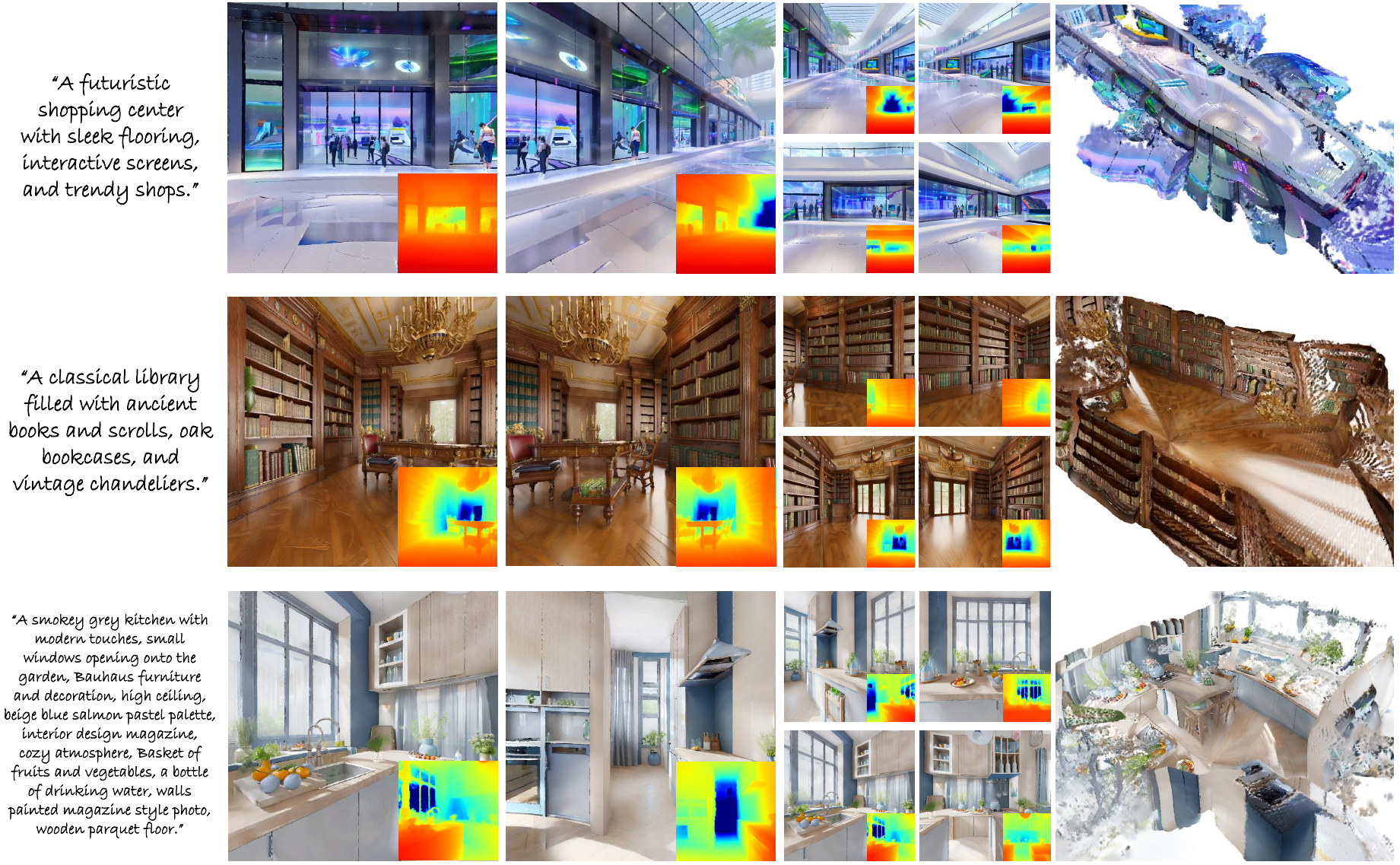}%
\label{results_indoor}}
\hfil
\subfloat[Generation Results of Outdoor Scenes.]{\includegraphics[width=0.98\textwidth]{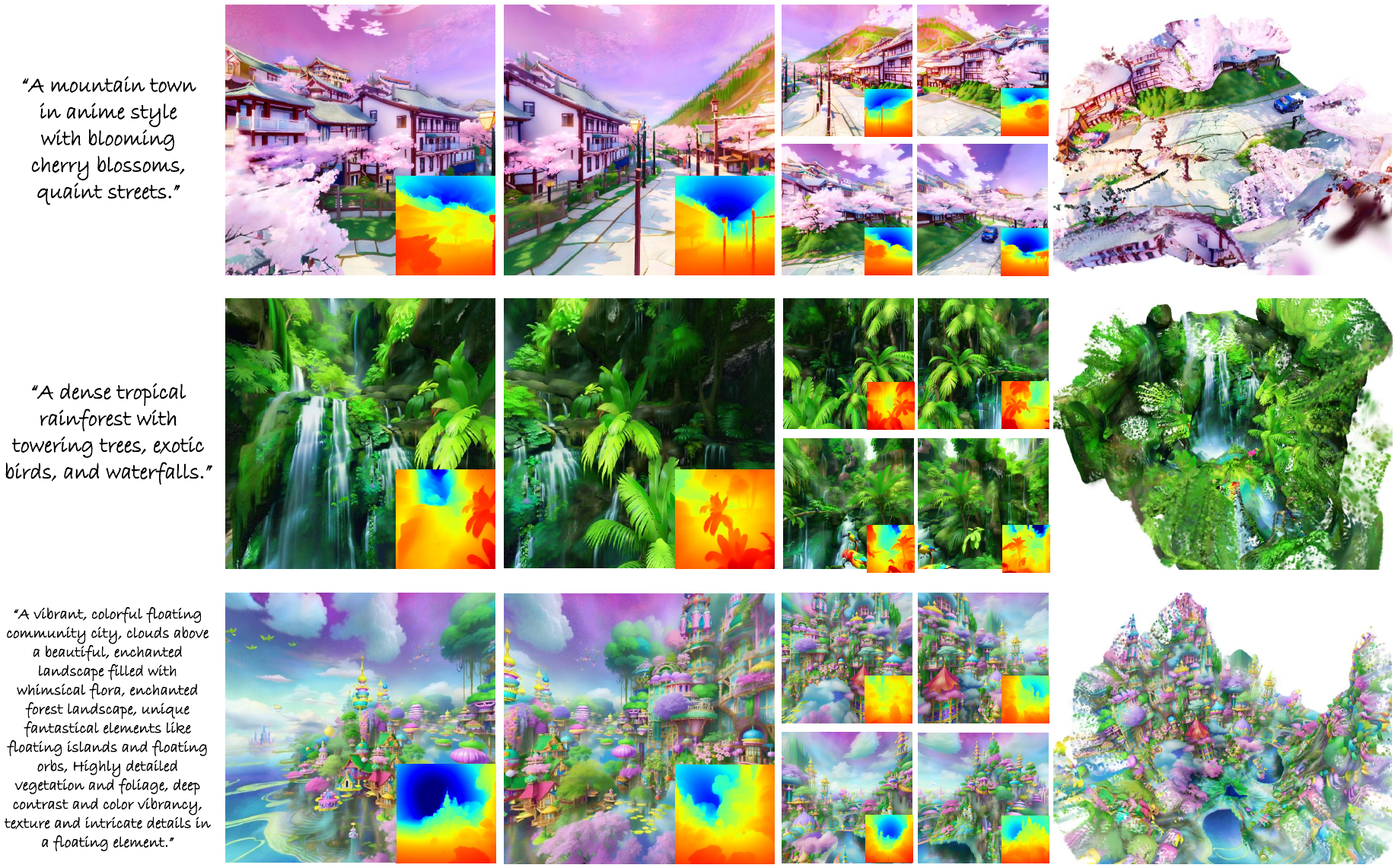}%
\label{results_outdoor}}
\caption{More results of our 3D scene generation. Our approach can generate fully enclosed 3D scenes with high consistency and style diversity, both for indoor and outdoor environments. }
\label{fig:more_results}
\end{figure*}

Our proposed HoloDreamer can generate highly view-consistent, immersive and fully enclosed 3D scenes based on text descriptions, as shown in Fig.~\ref{fig:results}. Furthermore, the pipeline exhibits a high degree of generality, encompassing a diverse spectrum of styles ranging from interior to exterior environments, as shown in Fig.~\ref{fig:more_results}.
Comprehensive experiments strongly demonstrate that our approach surpasses other text-driven 3D scene generation methods in terms of overall visual consistency and harmony, reconstruction quality, and rendering robustness when it comes to generating full-enclosed scenes. 

In summary, our contributions can be outlined as follows:

\begin{itemize}
	\item We propose HoloDreamer, a text-driven 3D scene generation approach that combines diffusion models and 3D Gaussian Splatting to generate fully enclosed immersive 3D scenes with visual consistency.

	\item We introduce Stylized Equirectangular Panorama Generation, a framework for panorama generation by leveraging the power of diffusion models, which can maintain the geometric features of equirectangular projection while expanding the range of generative capabilities and diversity.
 
	\item Our proposed Enhanced Two-Stage Panorama Reconstruction module provides multi-view constraints for the single viewpoint of the panorama and introduces inpainting to 3D-GS optimization, reducing artifacts and improving the integrity of the scene, achieving fast and high fidelity 3D reconstruction from single panorama.

\end{itemize}

\section{Related Works}
\subsection{3D Representation}
The field of 3D representation has seen a multitude of approaches, each with its own set of trade-offs and applications. Traditional primitives such as point clouds, meshes and voxels have been the cornerstone of 3D modeling for years. However, these methods face limitations in terms of representational ability: they often require a large quantity of data to achieve high resolution, which can be cumbersome and computationally expensive.
With the advent of deep learning, implicit neural representations have emerged as a powerful alternative, including Signed Distance Functions (SDF)\coloredcite{park2019deepsdf}, Occupancy Networks\coloredcite{mescheder2019occupancy} and Neural Radiance Fields (NeRF)\coloredcite{mildenhall2020nerf}. Especially, NeRF has been demonstrated the ability to represent complex 3D shapes and textures with rich details, and has been applied extensively. Nonetheless, these methods are not without their challenges. Implicit forms can be difficult to handle. What's more, training process is time-consuming, and often relies on a considerable number of views to optimize the representation, which may not always be feasible.

More recent advancements have led to the development of 3D Gaussian Splatting (3D-GS)\coloredcite{kerbl20233d} , a novel approach that offers a more efficient and versatile method for the representation of 3D scenes. This method could represent complete and unbounded 3D scenes by effectively `splatting' Gaussians. Spherical harmonics and opacity ensure strong representation capabilities, while differentiable rasterization greatly improves rendering speed and optimization efficiency. It can be initialized based on a point cloud which is widely applied in many scenarios and relatively easy to acquire, as a strong reference of initial positional and geometric information. In addition to this, the process involves a split-and-clone mechanism that could naturally propagate new Gaussians, allowing for gradual supplementation of intricate details.

Balancing the quality and efficiency of reconstruction, 3D-GS is our most suitable choice, and the split-and-clone process provides the foundation for inpainting in 3D scenes.

\subsection{3D Scene Generation}
3D content generation has become a focal point in the field of AI-generated content (AIGC). Generative Adversarial Networks (GAN)\coloredcite{goodfellow2020generative} was once particularly influential in 2D creation. Inspired by this, a range of GAN models are designed to produce 3D content within specific domains, such as faces, cars, cats and chairs. 3D-GAN\coloredcite{wu2016learning}, l-GAN\coloredcite{achlioptas2018learning}, and Tree-GAN\coloredcite{shu20193d} utilize simple explicit primitives to represent textureless geometric shapes. HoloGAN\coloredcite{nguyen2019hologan} and BlockGAN\coloredcite{nguyen2020blockgan}, on the other hand, learn geometric and textural representations to generate textured 3D content. GRAF\coloredcite{schwarz2020graf}, Pi-GAN\coloredcite{chan2021pi}, as well as Giraffe\coloredcite{niemeyer2021giraffe}, leverage implicit neural networks to achieve superior consistency and fidelity in the generated 3D scenes. However, training GANs is notoriously difficult due to their complex and unstable training dynamics. Beyond that, GANs struggle to effectively handle text prompts, leading to limited controllability, and their outputs are constrained by the specific training datasets, which prevents widespread application. There are some efforts, such as Point-E\coloredcite{nichol2022point} and Shape-E\coloredcite{jun2023shap}, that train more stable diffusion models\coloredcite{rombach2021highresolution} to generate 3D object end to end. But due to the scarcity of high-quality paired text-to-3D datasets, the generated content remains confined to specific domains and exhibits relatively coarse geometries and textures.

More recently, the emergence of language-image pre-trained models has catalyzed a multitude of zero-shot tasks and also has emerged as a potent tool in text-driven 3D generation. A significant amount of effort is invested in utilizing semantic priors in pre-trained models to generate domain-free objects and scenes with a high degree of detail and coherence. CLIP-NeRF\coloredcite{wang2022clip} and DreamFields\coloredcite{jain2022zero} use the priors of CLIP\coloredcite{radford2021learning} for supervision of optimization. 
Diffusion Models have made strides for generating complex data distributions\coloredcite{rombach2021highresolution,saharia2022photorealistic,ramesh2021zero,nichol2021glide}. DreamFusion\coloredcite{poole2023dreamfusion} introduces a method called Score Distillation Sampling (SDS), which distills high-level semantic priors from diffusion models to optimize 3D representations within different viewpoints, ensuring the consistency across viewpoints and correspondence between the prompt and the generated 3D objects. The techniques inspired a quantity of subsequent works, such as Magic3D\coloredcite{lin2023magic3d}, ProlificDreamer\coloredcite{wang2024prolificdreamer}, HiFi-123~\coloredcite{yu2023hifi}, Progressive3D~\coloredcite{cheng2023progressive3d} and DreamGaussian\coloredcite{tang2023dreamgaussian}. However, these methods only work when generating objects with relatively simple geometry, but are unable to generate large, wrap-around 3D scenes with rich details.

Several studies\coloredcite{fridman2024scenescape,hollein2023text2room, chung2023luciddreamer,zhang2024text2nerf} harness the low-level priors of diffusion models as direct and explicit supervision to generate large 3D scenes from text prompts. Initially, a 2D image is either provided by the user or generated from the text prompts using a diffusion model. Subsequently, a monocular depth estimation model is employed to infer the corresponding depth information, thereby transferring the 2D image information into a 3D context. To cover a vast field of view and generate large-scale 3D scenes, these methods progressively apply the diffusion model to outpaint existing scene, following a preset trajectory. Early endeavors utilized mesh as the 3D representation. SceneScape\coloredcite{fridman2024scenescape} generates scenes that recede into the distance through a zoom-out trajectory, while Text2Room\coloredcite{hollein2023text2room} is primarily confined to indoor scenes. However, the capabilities of mesh as a 3D representation are quite limited. More recent approaches, such as LucidDreamer\coloredcite{chung2023luciddreamer} and Text2NeRF\coloredcite{zhang2024text2nerf}, capitalize on the robust and flexible 3D representational capabilities of 3D-GS\coloredcite{kerbl20233d} and NeRF\coloredcite{mildenhall2020nerf} to synthesize 3D scenes that are not restricted to specific domains.
However, each outpainting iteration only perceives a fraction of the existing scene, not the entire scene. This can lead to chaotic objects and overall visual inconsistencies. Additionally, the camera pose of each image outpainting step cannot be effectively constrained by diffusion models, resulting in viewpoints in the preset trajectory that are often nearly horizontal, and instability often occurs when generating the top and bottom parts. To address these limitations, our approach generates a panorama from text prompts that directly covers the panoramic field of view, followed by a 3D-GS reconstruction from the single panorama. This methodology significantly enhances the visual consistency of the scene and enables the generation of diverse fully enclosed 3D scenes.

\subsection{Panorama Generation}
Panorama has a wide and unobstructed view that catches a vast area of scene. Some works, such as PanoGen\coloredcite{li2024panogen} and MultiDiffusion\coloredcite{bar2023multidiffusion}, utilize pre-trained diffusion models to generate long-image from text prompt. However, these so-called ``panoramas'' are essentially stitched from a series of perspective images. They do not align with the true projection relationships inherent in panorama, and there is an absence of view-consistency across the entirety. Additionally, this kind of panoramas don't include a full 360-degree horizontal field of view. Furthermore, there is often discontinuity between the leftmost and rightmost parts of the image. 

An intuitive solution to generate 360-degree panoramas that conforms to the equirectangular projection involves fine-tuning models using 360-degree panorama database. MVDiffusion\coloredcite{tang2023MVDiffusion} introduces a Correspondence-aware Attention (CAA) mechanism to simultaneously denoise and generate eight images that are consistent across views. However, all eight images are in the vertically middle area, which means it falls short of generating images covering the top and bottom parts of the panorama.
StitchDiffusion\coloredcite{wang2024customizing}, on the other hand, performs LoRA\coloredcite{hu2021lora} fine-tuning to generate the whole 360-degree panorama and employs a global crop to ensure continuity between the leftmost and rightmost parts of the image. While Diffusion360\coloredcite{feng2023diffusion360} uses dreambooth\coloredcite{ruiz2023dreambooth} fine-tuning and utilizes circular blending techniques to prevent discontinuities.
For our approach, we employ the pre-trained Diffusion360 model to generate the base panorama that highly conforms to the equirectangular projection.

\begin{figure*}[!t]
\centering
\subfloat[Stylized Equirectangular Panorama Generation]{\includegraphics[width=0.59\linewidth]{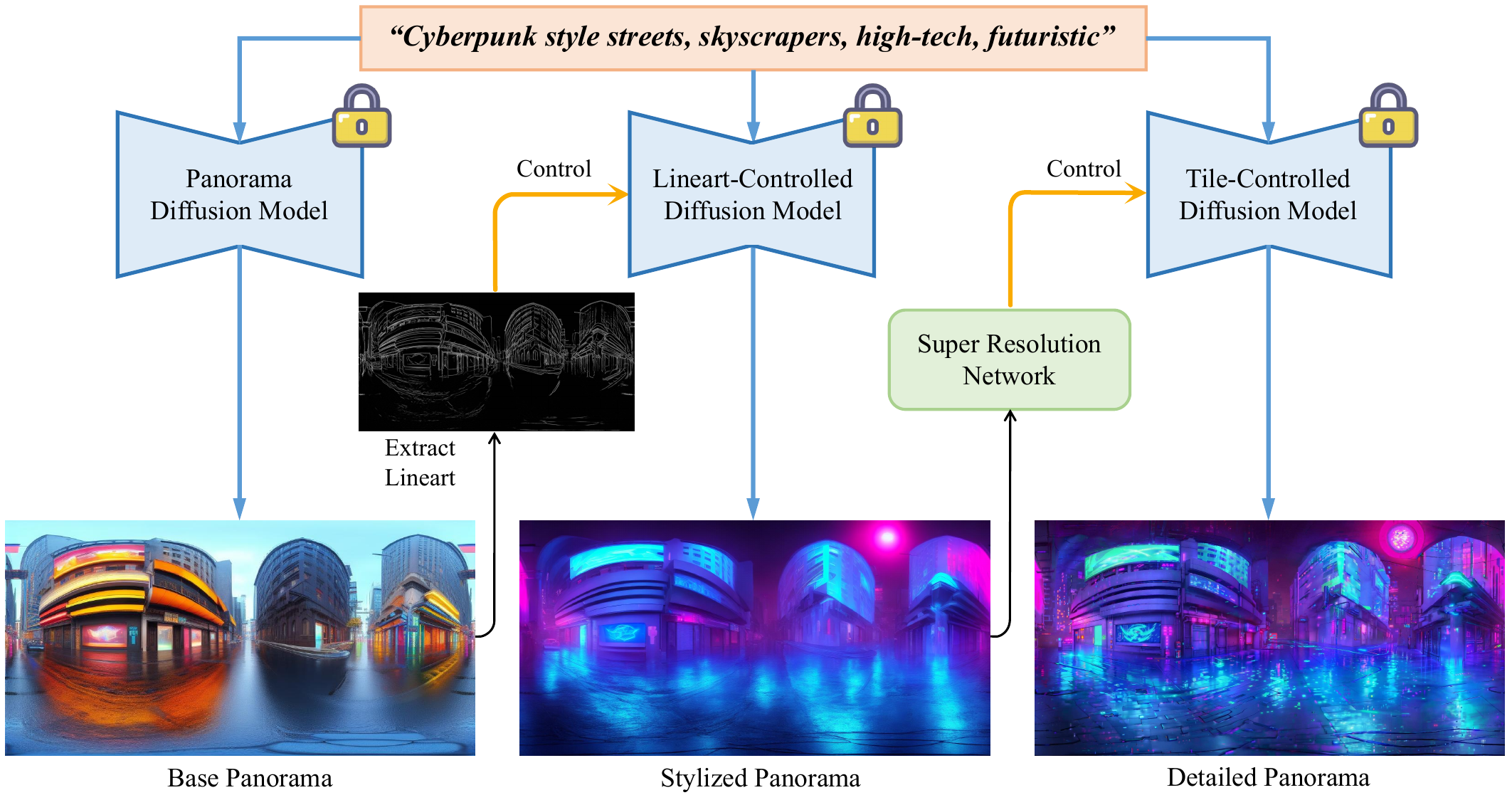}%
\label{pipe}}
\hfil
\subfloat[Effectiveness of Circular Blending]{\includegraphics[width=0.38\linewidth]{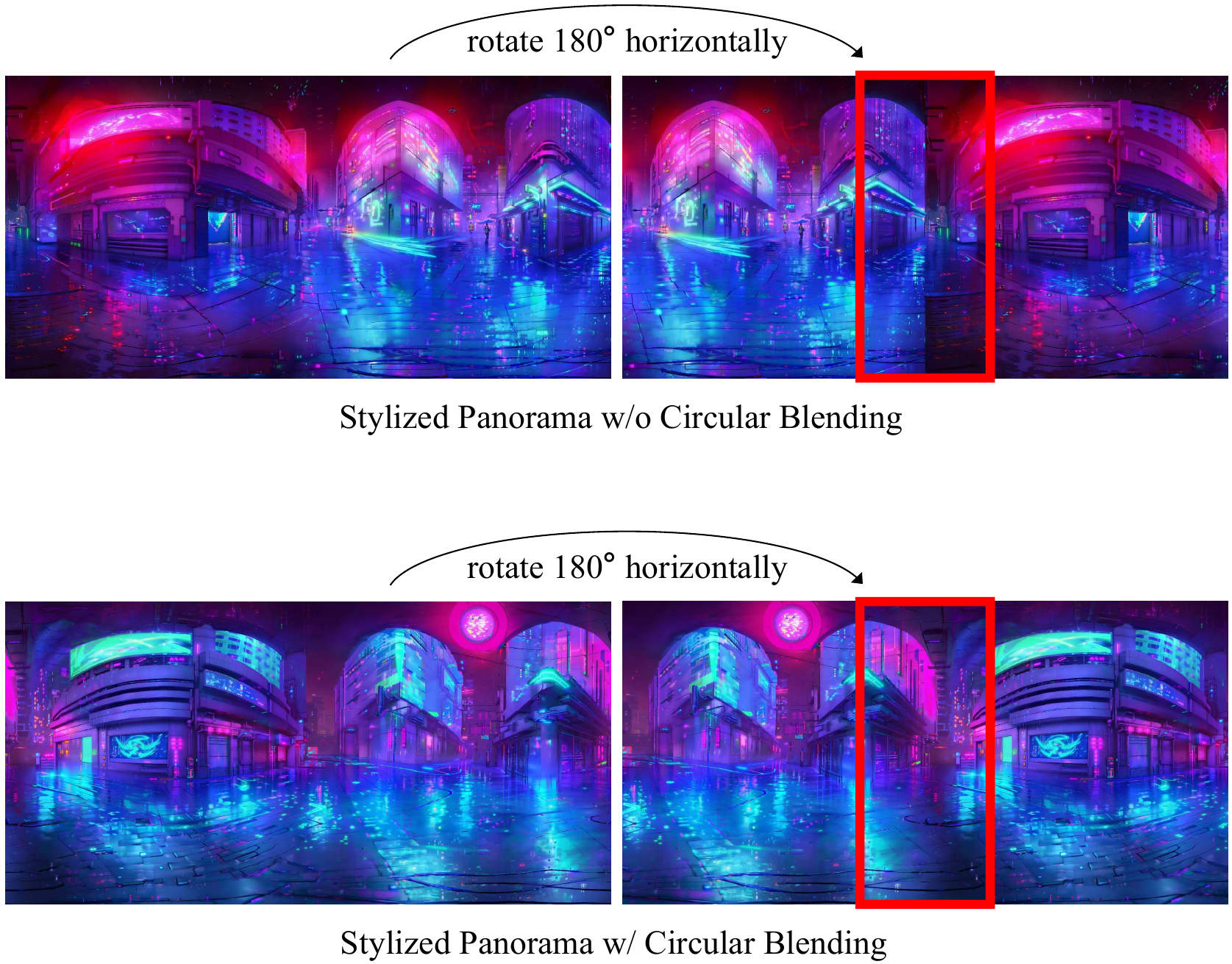}%
\label{blending}}
\caption{Overview of our Stylized Equirectangular Panorama Generation. Given a user prompt, multiple diffusion models are used to generate stylized high-quality panoramas. Additionally the circular blending technique is applied to avoid cracks when rotating the panorama.}
\label{fig:gen_pipe}
\end{figure*}

\section{Method}
\begin{figure*}[!t]
\centering
\includegraphics[width=\textwidth]{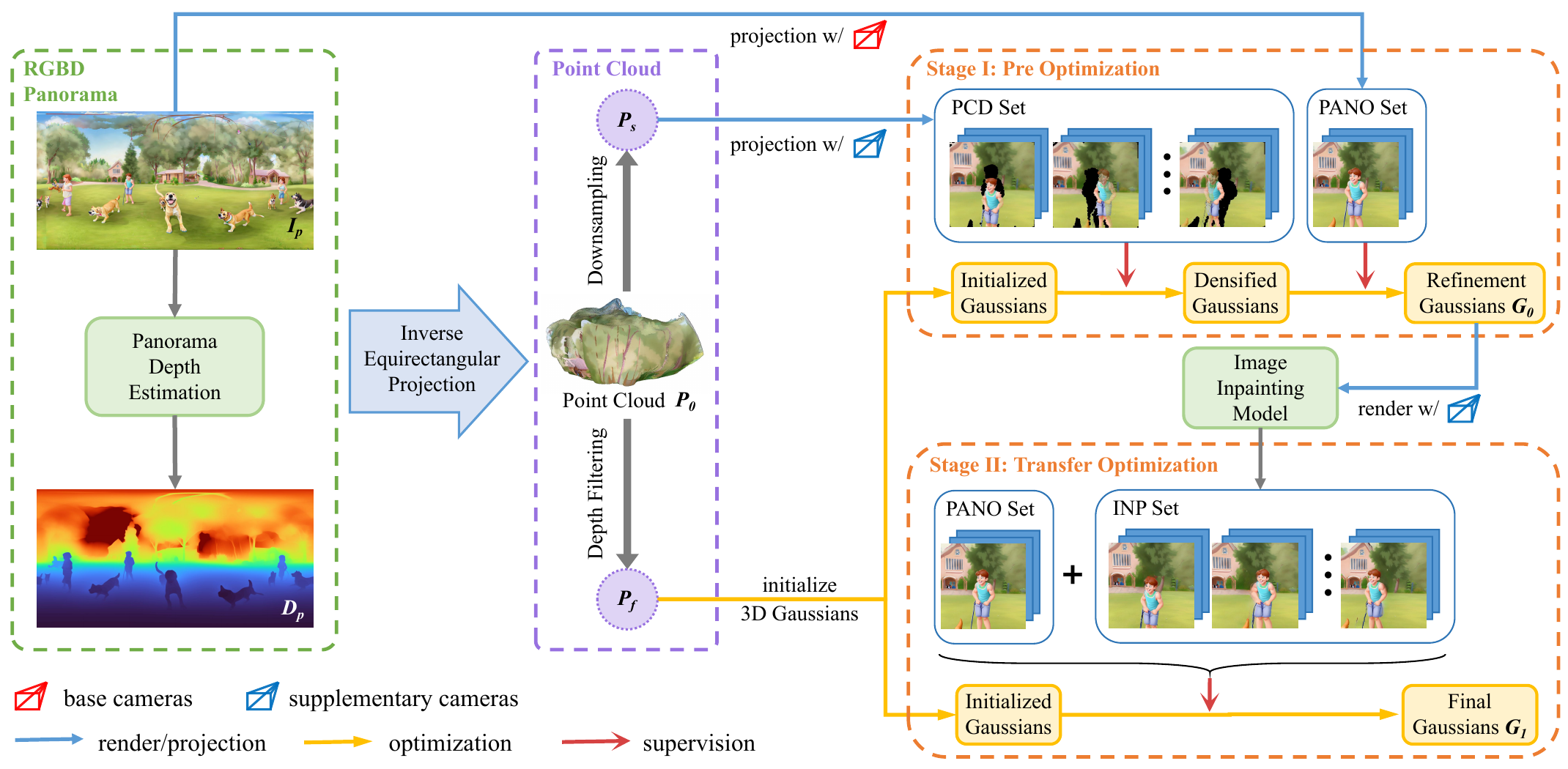}
\caption{Overview of our Enhanced Two-Stage
Panorama Reconstruction. We perform depth estimation on the generated panorama and then project RGBD data to obtain the point cloud. Two types of cameras --- base cameras and supplementary cameras --- for projection and rendering in different scenarios, and prepare three image sets for supervision at different stages of 3D-GS optimization. The rendering images of the reconstructed scene from Pre Optimization stage are inpainted for optimization in Transfer Optimization stage, resulting in the final reconstructed scene.}
\label{fig:reco_pipe}
\end{figure*}

We propose a text-driven 3D scene generation framework that is capable of generating fully enclosed immersive scenes with a high level of overall visual effect and rendering robustness. Firstly, we use the diffusion model to progressively generate stylized, high-quality equirectangular panorama with high view harmony based on text prompts, as shown in Fig.~\ref{fig:gen_pipe}, which is introduced in the following Sec.~\ref{subsec:gen}. And then we perform two-stage panorama reconstruction represented by 3D-GS with enhanced integrity, as shown in Fig.~\ref{fig:reco_pipe}, which is introduced in the following Sec.~\ref{subsec:recon}. 

\definecolor{myred}{RGB}{255,0,0}
\definecolor{myblue}{RGB}{0,112,192}

\subsection{Stylized Equirectangular Panorama Generation} \label{subsec:gen}
In order to generate a panorama that geometrically conforms to the equirectangular projection, we apply a diffusion model that is fine-tuned using a comprehensive panorama database including both indoor and outdoor scenes. This fine-tuning process enables the model to adeptly capture the geometric principles and distinctive features inherent to the equirectangular projection. The model is used to generate the initial base panorama with a relatively reduced resolution. Its geometric features of an equirectangular projection set the groundwork for subsequent refinements and enhancements. 

During the denoising process, we integrate a circular blending technique in the Diffusion360\coloredcite{feng2023diffusion360} framework, facilitating the creation of a seamless panorama. In the inference phase, following each sampling iteration, the left border and the right border of the image in the latent space are subjected to a blending procedure. This method is meticulously designed to maintain spatial continuity across left and right boundaries. We extend the application of this technique to apply it on all diffusion models discussed within this subsection, thereby effectively preventing the emergence of cracks when rotating the panorama, as shown in Fig.~\ref{fig:gen_pipe}\subref{blending}.

Given that the majority of existing panorama datasets are comprised of real-world images, there is a risk of overfitting when employing fine-tuning techniques. This leads to a resulting domain that is highly constrained for the generated data. To mitigate this issue, we first extracted the lineart from the generated base panorama. The form of lineart effectively retains the geometric features of the equirectangular projection. Subsequently, we use a lineart-controlled diffusion model with extracted lineart as a conditional control to generate stylized panorama endowing the panorama with the ability to express to express a diversity of novel styles and features that extend beyond the confines of the original database, without compromising the geometric characteristics.

To ensure the clarity and visual appeal of the reconstructed scene in the forthcoming panorama reconstruction, it is essential to procure a panorama of superior resolution and enhanced detail. To this end, we harness a super-resolution network to elevate the resolution of the stylized panorama. Subsequently, we apply a tile-controlled diffusion model, which serves to augment the details of the image. This meticulous process culminates in the derivation of our final detailed panorama that will be utilized in the subsequent subsection.

\subsection{Enhanced Two-Stage Panorama Reconstruction} \label{subsec:recon}
\noindent {\bf{Depth Estimation.}}
Considering the diverse distributions of generated panoramas, our panorama depth estimation methodology must have a high degree of generalization ability, applicable to both bounded indoor scenes and unbounded outdoor scenes. We apply 360MonoDepth\coloredcite{rey2022360monodepth}, a zero-shot framework, which utilizes a pre-trained monocular depth estimation model to achieve high-resolution panorama depth estimation. 
The process involves projecting the panorama onto twenty perspective tangent images, each corresponding to a face of an icosahedron. Using state-of-the-art monocular depth estimation techniques, we ascertain the disparity of each individual image. Subsequently, these disparity maps are globally aligned and blended into the panprama's disparity map. We opt for the frustum blending method to combine the images, thereby effectively enhancing global smoothness.

After blending, we obtain the overall disparity map of the generated panorama. However, the scale and offset of the map remain ambiguous and require calibration to ensure accuracy. To convert the disparity map to an absolute depth map, we randomly select a subset of perspective tangent faces from the icosahedron. Utilizing a pre-trained metric depth estimation model, we estimate absolute depth on the images corresponding to the chosen faces, and then convert these obtained metric depth maps into disparity maps, which serve as the ground truth with reference scales and offsets. By minimizing the sum of squared differences between the overall disparity map of the generated panorama and the set of reference perspective disparity maps, we determine the parameters --- global offset and scale --- using the least squares method. finally obtaining the metric depth of the generated panorama.

\begin{figure}[!t]
\centering
\subfloat[Raw]{\includegraphics[width=0.33\linewidth]{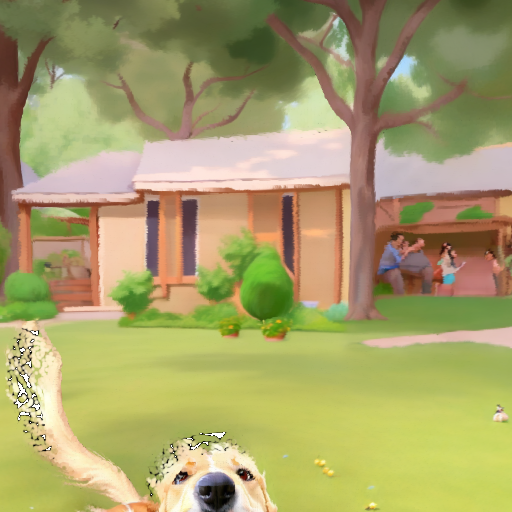}%
\label{raw_pcd}}
\hfill
\subfloat[Depth Filtered]{\includegraphics[width=0.33\linewidth]{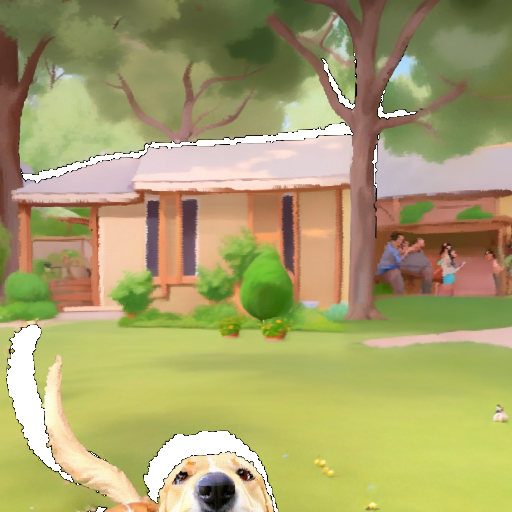}%
\label{fltr_pcd}}
\hfill
\subfloat[Downsampled]{\includegraphics[width=0.33\linewidth]{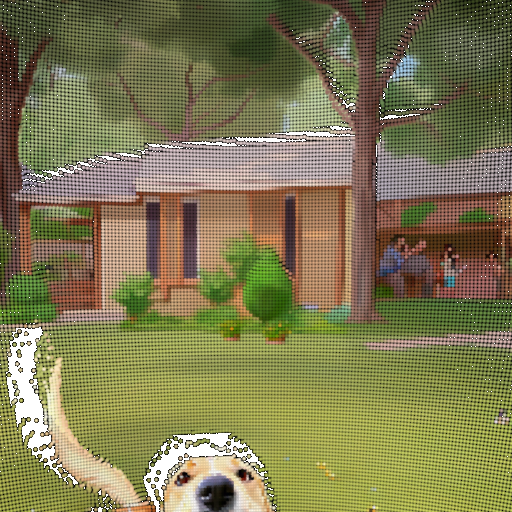}%
\label{ds_pcd}}
\caption{Visual comparison of point cloud data: raw point cloud $\boldsymbol{P_0}$, depth filtered point cloud $\boldsymbol{P_f}$, and downsampled point cloud $\boldsymbol{P_s}$.}
\label{fig:pcd}
\end{figure}

\noindent {\bf{{Point Cloud Reconstruction.}}}
Given the RGB image $\boldsymbol{I_p}$ and corresponding depth map $\boldsymbol{D_p}$ of the panorama, data in the form of point cloud can be easily obtained. We conduct a reverse equirectangular projection from the RGBD panorama image onto a raw point cloud, denoted as $\boldsymbol{P_0}$. The projection converts the pixel coordinates of each pixel of the panorama into 3D world coordinates by determining the corresponding longitude and latitude. In this context, the camera position of the panorama is fixed as the center of the sphere, which also serves as the origin of the world coordinate system. The reverse projection is represented by the following formula:
\begin{equation}
\label{deqn_ex1}
\boldsymbol{P_0} = \phi^{-1}_{erp}([\boldsymbol{I_p}, \boldsymbol{D_p}]).
\end{equation}
However, a prevalent issue across almost all existing depth estimation models is the depth mixing problem\coloredcite{hu2022deep}, which manifests as a difficulty in accurately discerning the depth of pixels proximal to object boundaries. This challenge often leads to a blending of depth values at the edges, thereby introducing artifacts in both the raw point cloud and the subsequent reconstructed 3D-GS. Despite this, the depth gradient at the edges tends to be notably steep compared to other areas of the image. To mitigate the problem, we introduce a preprocessing step that involves calculating the 2D gradient on the depth map of the panorama. Subsequently, we apply a threshold-based filter to exclude points with excessively high gradients, resulting in a point cloud denoted as $\boldsymbol{P_f}$ that is more free from artifacts, which is utilized for the initialization of the 3D-GS, as shown in Fig.~\ref{fig:pcd}. The initialization provides a high degree of depth and geometric priors. What's more, we avoid resetting the opacity during the 3D-GS optimization to avoid losing the spatial information obtained from the point cloud.

\noindent {\bf{{Two-stage 3D-GS Reconstruction.}}}
To effectively utilize information of multiple formats, we design two types of cameras, base cameras and supplementary cameras, for projection and rendering in different scenarios, as well as three distinct image sets derived from different types of data, named PCD set, PANO set, and INP set, for supervision across various stages of optimization. This strategy helps us achieve optimal performance in both reconstruction speed and quality, and significantly enhance rendering robustness.

Specifically, the panorama $\boldsymbol{I_{p}}$ is converted to a series of perspective images for supervision of the 3D-GS optimization. We configure a set of $M$ base cameras, with each camera sharing the same intrinsic parameters denoted by $\boldsymbol{K}$. The extrinsic parameters are denoted as $\boldsymbol{E_i}$, describing the specific pose of the $i$-th base camera. These camera poses are strategically arranged to provide coverage across the entirety of a sphere projected from the panorama. All cameras are positioned at the center of the sphere. These images projected from panorama $\boldsymbol{I_{p}}$ using base cameras, constitute the PANO set and inherit the high-resolution characteristics of the panorama $\boldsymbol{I_{p}}$. Supervision with PANO set ensures the fidelity of the reconstruction. The projection is based on the following formula:
\begin{equation}
\label{deqn_ex2}
\boldsymbol{I_i} = \phi_{erp2pers}(\boldsymbol{I_p}, \boldsymbol{K}, \boldsymbol{E_i}).
\end{equation}
Where $\boldsymbol{I_i}$ is the image projected by the $i$-th base camera.

\begin{figure}[!t]
\centering
\includegraphics[width=\linewidth]{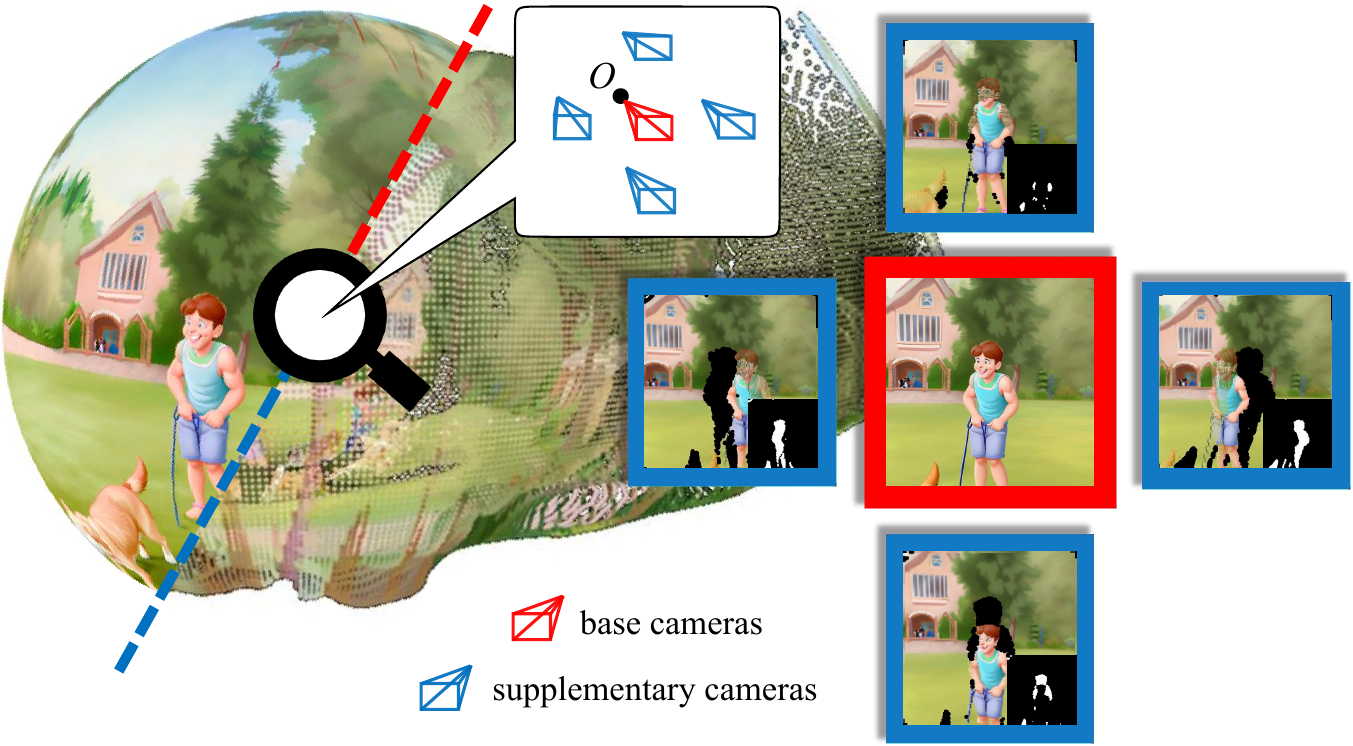}%
\caption{The relationship between the base camera and corresponding supplementary cameras as well as their projection results. Supplementary cameras surround the corresponding base camera, each with variations in position and orientation. The base camera is used for projection of the panorama, while the supplementary cameras are used for projecting point clouds.}
\label{fig:cam_proj}
\end{figure}

However, the images in PANO set offer very limited camera poses. Because it is designed for application to panoramic images, the position of the base cameras is restricted to a single location. The scarcity of diverse viewpoints for supervision can easily lead to overfitting on constrained poses and poor rendering robustness when the camera moves, including the emergence of visual artifacts and excessively elongated 3D Gaussians. To overcome this,
we add an additional sample of $N$ supplementary cameras with the same shared intrinsic parameters $\boldsymbol{K}$ and extrinsics $\boldsymbol{E_{ij}}$ which surround the corresponding base camera with extrinsics $\boldsymbol{E_i}$, where $j$ ranges from $1$ to $N$. Their positions and orientations have both changed compared to the corresponding base camera as shown in Fig.~\ref{fig:cam_proj}, offering multi-view supplementation. For each supplementary view, we obtain the images that constitute PCD set by projecting point clouds. Considering the efficiency of projection, the original point cloud $\boldsymbol{P_0}$ is first downsampled to lower the density, as shown in Fig.~\ref{fig:pcd}. The downsampled point cloud $\boldsymbol{P_s}$ is projected from the world coordinate system to the pixel coordinate system using the following formula: 
\begin{equation}
\label{deqn_ex3}
\boldsymbol{I_{ij}}, \boldsymbol{M_{ij}} = \phi_{3\rightarrow2}(\boldsymbol{P_s}, \boldsymbol{K}, \boldsymbol{E_{ij}}).
\end{equation}
Where $\boldsymbol{I_{ij}}$ is the perspective image projected by the $j$-th supplementary camera of the $i$-th base camera and $\boldsymbol{M_{ij}}$ is the corresponding mask that illustrates the missing areas in the supplementary view $\boldsymbol{E_{ij}}$, which will be filled in to improve the integrity during the subsequent 3D-GS optimization process.

Initiating with the filtered point cloud $\boldsymbol{P_f}$, we execute a two-stage 3D-GS optimization utilizing the previously derived image sets. In the Pre Optimization stage, the PCD set is initially engaged for supervision. These images furnish multi-view constraints within the initial densification process, ensuring the appropriate spatial arrangement of the emergent 3D Gaussians. Nonetheless, the downsampling process of the point cloud diminishes the resolution of images within the PCD set, leading to a loss of clarity in the preliminary Gaussian representations. To restore more details in the panorama, the PANO set, characterized by its high-resolution images derived from the panorama, is then used solely for supervision to refine the densified Gaussians. This subsequent refinement of the 3D Gaussians is instrumental in preserving the fidelity of the reconstructed scene $\boldsymbol{G_{0}}$.

Because the panorama has only a single viewpoint, the reconstructed scene contains numerous missing regions as a result of object occlusions. In the second stage, we render Gaussians $\boldsymbol{G_{0}}$ for each supplementary view $\boldsymbol{E_{ij}}$ and then use an image inpainting model to obtain $\boldsymbol{V_{ij}}$ filled in the missing pixels. The formula is as follows:
\begin{equation}
\label{deqn_ex3}
\boldsymbol{V_{ij}}, = F_{inpaint}(R_G(\boldsymbol{G_0}, \boldsymbol{K}, \boldsymbol{E_{ij}}), \boldsymbol{M_{ij}}).
\end{equation}
Where $i$ ranges from $1$ to $M$, and $j$ ranges from $1$ to $N$, these inpainted images constitute the third image set, named INP set.
Ultimately, we perform Transfer Optimization on newly initialized Gaussians. The INP set and the PANO set are integrated as supervision concurrently to achieve final reconstruction Gaussians $\boldsymbol{G_1}$ with greater rendering robustness. The split-and-clone process of 3D-GS automatically inpaint missing regions during optimization.

\section{Experiments}
In this section, we employ rigorous and comprehensive experiments to demonstrate the superiority of our approach. The evaluation is segmented into two primary components: experiments of panorama generation in Sec.~\ref{subsec:exp_gen} and experiments of panorama reconstruction in Sec.~\ref{subsec:exp_recon}. We carefully compare our method with benchmark methodologies and conduct additional ablation study to evaluate the generation capability and the reconstruction performance of our framework, respectively.

\begin{figure*}[!t]
\centering
\includegraphics[width=0.98\textwidth]{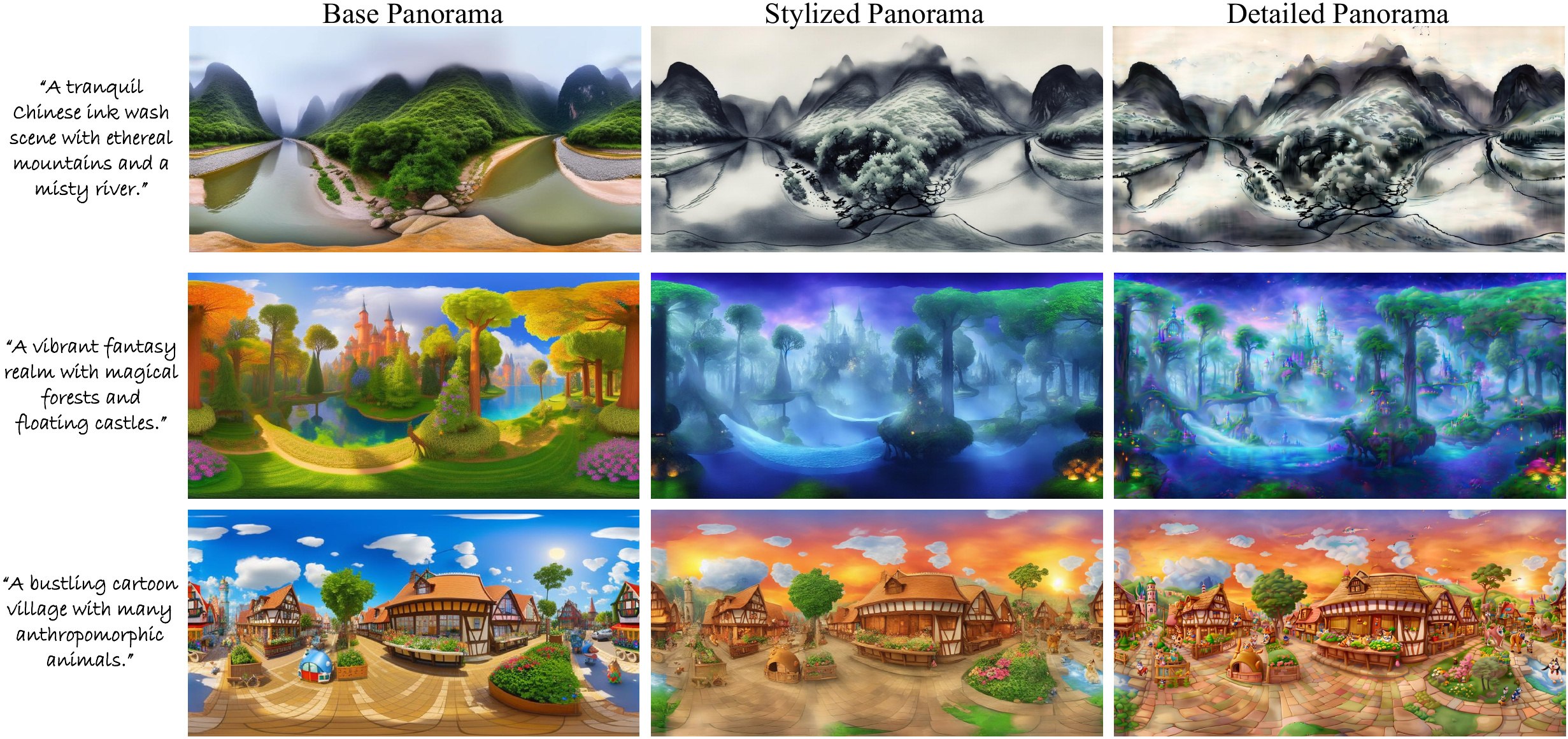}
\caption{Comparison of Base Panorama, Stylized Panorama, and Detailed Panorama. Stylized Panorama has a style that closely matches the description than Base Panorama. Furthermore, Detailed Panorama is added more details.}
\label{fig:gen_ablation}
\end{figure*}

\begin{figure*}[!t]
\centering
\includegraphics[width=0.94\textwidth]{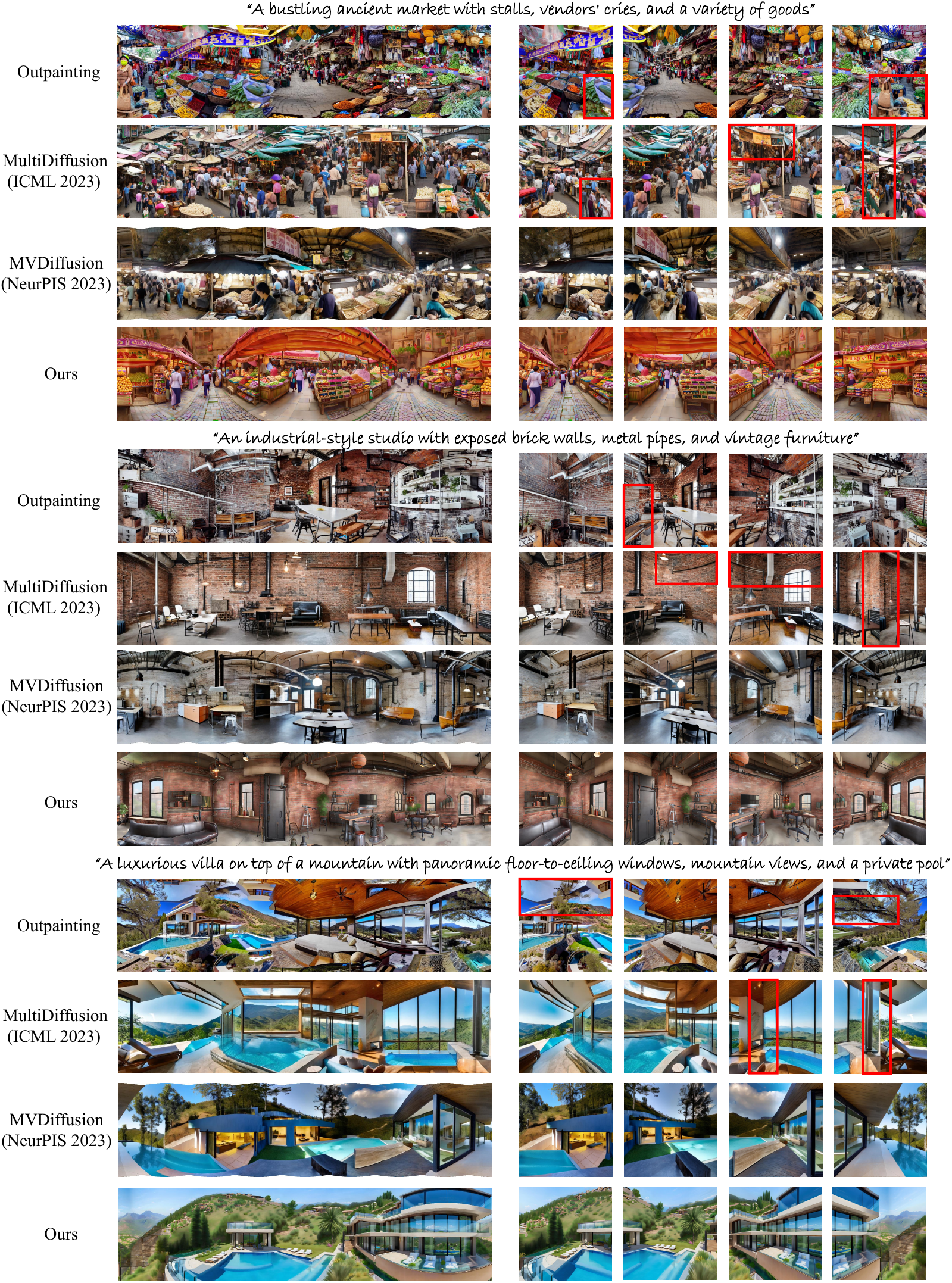}
\caption{Qualitative comparison of panoramas generated by our method and baselines based on diverse text prompts. We show the panoramas and middle faces of the corresponding cubemaps for an intuitive assessment of visual effects.}
\label{fig:gen_compare}
\end{figure*}

\subsection{Experiments of Panorama Generation}\label{subsec:exp_gen}
\noindent {\bf{Baseline Methods.}} 
We compare our Stylized Equirectangular Panorama Generation approach with three panorama generation methods: Outpainting, MultiDiffusion\coloredcite{bar2023multidiffusion} and MVDiffusion\coloredcite{tang2023MVDiffusion}. Outpainting is a widely utilized progressive generation method prevalent in the field of 3D scene generation. It initiates from an initial image and progressively generates outward extensions, guided by a reference diffusion model and following a preset camera trajectory. MultiDiffusion leverages a reference diffusion model to incrementally generate a long image that can be considered as the middle part of the 360-degree panorama. It achieves this by simultaneously constraining multiple image crops during the denoising process. Both methods require no training of the reference diffusion model. While MVDiffusion fine-tunes a diffusion model with correspondence-aware attention (CAA) mechanism on a panorama dataset enabling the generation of eight cross-view consistent images, but it encounters challenges in producing the top and bottom parts of the panorama. 

\begin{table}
\setlength{\tabcolsep}{2.5pt}
\caption{Quantitative Comparison of Image Aesthetic and Quality for Panoramas Generated by Our Method and Baselines in the Form of Panorama and Four Middle Faces of Cubemap.(\textbf{Best})\label{tab:gen_baseline}}
        \renewcommand{\arraystretch}{1.2}
	\centering
	\begin{tabular}{ccccc}
		\hline
            \multirow{2}*{Method} & 
                \multicolumn{2}{c}{Panorama} &
                \multicolumn{2}{c}{Cube Faces} \\
            \cline{2-5}
		~ & TANet\textuparrow & CLIP-Aesthetic\textuparrow & TANet\textuparrow & CLIP-Aesthetic\textuparrow \\
		\hline
            Outpainting & 5.397 & 5.755 & 5.438 & 5.739 \\
            MultiDiffusion\coloredcite{bar2023multidiffusion} & 5.459 & 5.956 & 5.392 & 5.848 \\
            MVDiffusion\coloredcite{tang2023MVDiffusion} & 5.394 & 5.642 & 5.463 & 5.775 \\
            \bf{Ours} & \bf{5.583} & \bf{6.198} & \bf{5.521} & \bf{6.017} \\
            \hline
	\end{tabular}
\end{table}

\noindent {\bf{Comparisons.}}
We compare our panorama generation approach with baseline methods on different text prompts in the form of both panorama and cubemap, as shown in Fig.~\ref{fig:gen_compare}. Due to the lack of global consideration, Outpainting can lead to severely chaotic objects and global inconsistencies. The panoramas generated by MultiDiffusion are planar and do not conform to equirectangular projection, resulting in distortion after projection onto perspective images, and failing to ensure 360-degree continuity. MVDiffusion generates multiple images and stitches them together, which reduces the overall consistency and harmony of the stitched panorama. In contrast, our method directly generates high-quality panoramas that conform to the equirectangular projection, achieving a high level of overall consistency and harmony in the scene, and also ensuring excellent visual effects in perspective views.

We employ two image aesthetic quality assessment metrics, TANet\coloredcite{herethinking} and CLIP-aesthetic\coloredcite{schuhmann2022laionb}, to quantitatively compare the aesthetic quality both on panoramic images and cubemap images, as shown in Tab.~\ref{tab:gen_baseline}. Our method achieved higher aesthetic quality scores on both forms of images, which demonstrates the superiority of our method over the baseline in terms of visual effects. Additionally, we compare the panoramas at different stages of the generation process in Fig.~\ref{fig:gen_ablation}, which confirms the effectiveness of applying multiple diffusion models in enhancing style and detail.

\begin{figure*}[!t]
\centering
\includegraphics[width=\linewidth]{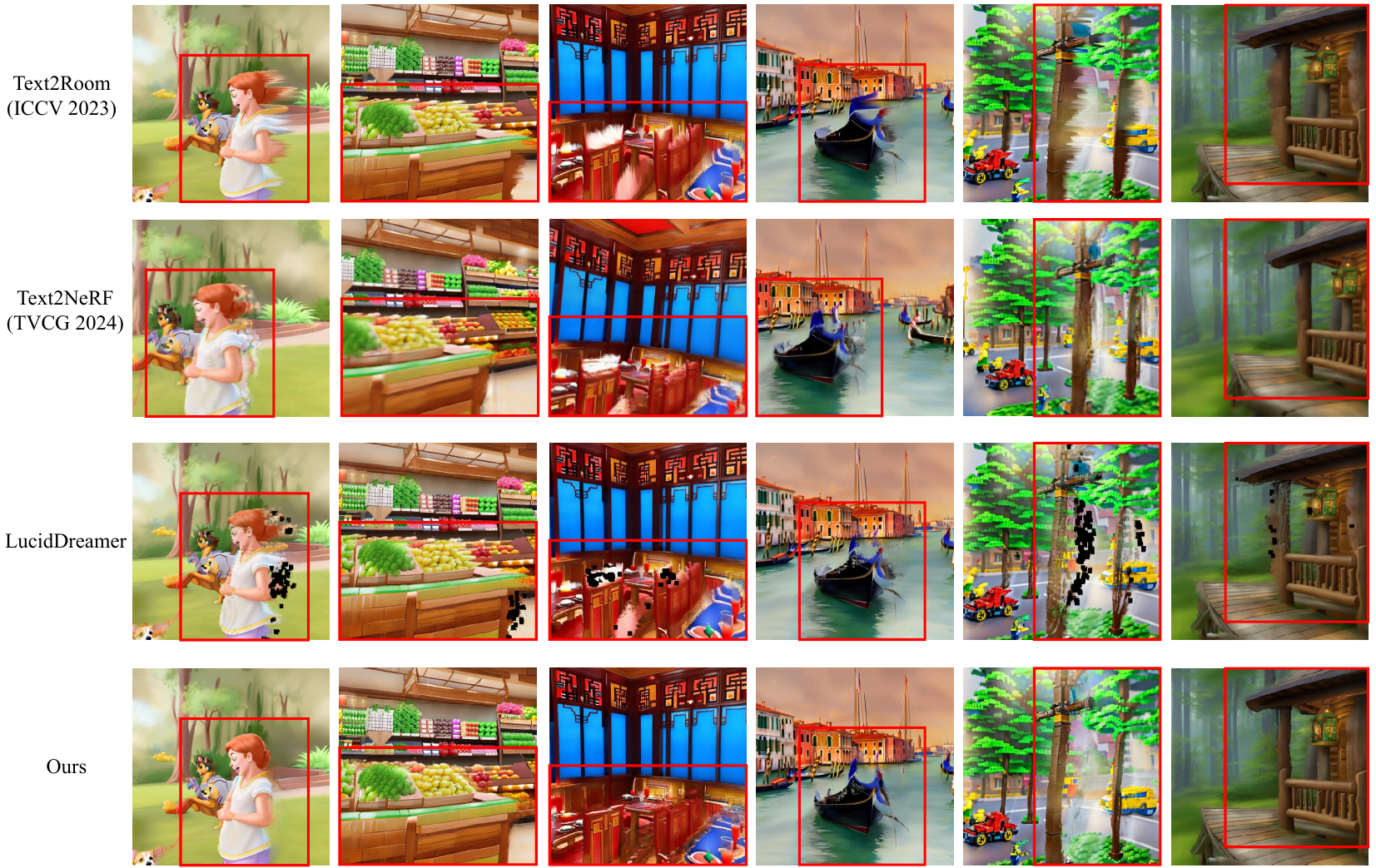}
\caption{Qualitative comparison of panorama reconstruction with baselines from different panoramas. Our method effectively avoids artifacts and fill in the missing areas, achieving better rendering robustness.}
\label{fig:recon_compare}
\end{figure*}

\subsection{Experiments of Panorama Reconstruction}\label{subsec:exp_recon}

\noindent {\bf{Baseline Methods.}} We compare our Enhanced Two-Stage Panorama Reconstruction approach with three 3D scene generation methods that combine the outpainting technique with different 3D representations: Text2Room\coloredcite{hollein2023text2room} represented with mesh, Text2NeRF\coloredcite{zhang2024text2nerf} represented with NeRF and LucidDreamer\coloredcite{chung2023luciddreamer} represented with 3D-GS. All these baseline methods utilize a diffusion model to outpaint rendered RGB image and progressively generate the overall scene. Text2Room directly extracts mesh from inpainted RGBD image to represent watertight indoor scenes. Text2NeRF uses inpainted RGBD images as supervision to train a NeRF network with proposed depth loss. While LucidDreamer projects the outpainted RGBD images into the point cloud, and subsequently projects multiple images from the point cloud for supervision during the optimization of 3D-GS. Given the RGB panorama images and corresponding depth map obtained by our method, we avoid outpainting and instead adjust the respective baseline methods to directly reconstruct the single panorama. For Text2Room, we extract the mesh directly from the panorama without filtering the mesh to avoid excessive holes. Text2NeRF is supervised using the PANO set projected from the panorama. And LucidDreamer is supervised using the images projected from raw point cloud $\boldsymbol{P_0}$.

\begin{figure}[!t]
\centering
\includegraphics[width=\linewidth]{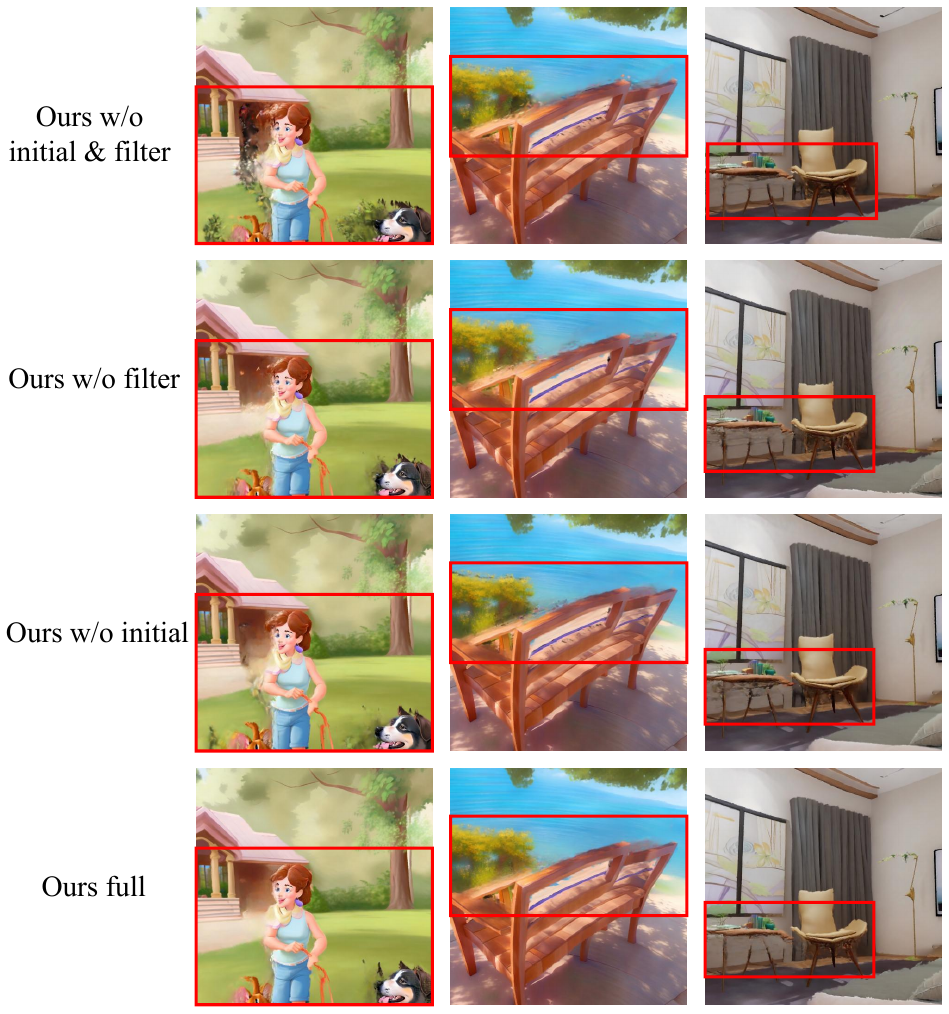}
\caption{Effective validation of depth filtering of point cloud and initial optimization with PCD set in our panorama reconstruction approach. These two components effectively mitigated the occurrence of artifacts.}
\label{fig:recon_ablation}
\end{figure}

\begin{table*}
\caption{Quantitative Comparisons of Panorama Reconstruction from Single Panorama on 3D Representation, Reconstruction Quality, Rendering Robustness and Reconstruction Time. (\textbf{Best})\label{tab:recon_baseline}}
        \renewcommand{\arraystretch}{1.2}
	\centering
	\begin{tabular}{cccccccc}
		\hline
            \multirow{2}*{Method} & \multirow{2}*{3D Representation} & 
                \multicolumn{3}{c}{Reconstruction Quality} &
                \multicolumn{2}{c}{Rendering Robustness} & 
                \multirow{2}*{Reconstruction Time (hours)} \\
            \cline{3-7}
		~ & ~ & PSNR\textuparrow & SSIM\textuparrow & LPIPS\textdownarrow & NIQE\textdownarrow & BRISQUE\textdownarrow & ~ \\
		\hline
            Text2Room\coloredcite{hollein2023text2room} & mesh & 34.497 & 0.957 & \bf{0.036} & 5.935 & 32.829 & \bf{0.008} \\
            Text2NeRF\coloredcite{zhang2024text2nerf} & NeRF & 34.336 & 0.927 & 0.162 & 7.316 & 32.891 & 9.738 \\
            LucidDreamer\coloredcite{chung2023luciddreamer} & 3D-GS & 34.501 & 0.958 & 0.068 & 6.255 & 44.738 & 0.962 \\
            \bf{Ours} & 3D-GS & \bf{40.189} & \bf{0.984} & 0.041 & \bf{5.372} & \bf{32.372} & 0.271 \\
            \hline
	\end{tabular}
\end{table*}

\noindent {\bf{Comparisons.}}
We compare our panorama reconstruction approach with baseline methods on different generated panoramas, as shown in Tab.~\ref{tab:recon_baseline}. We evaluate the reconstruction quality, rendering robustness and average reconstruction time of the panorama reconstruction. For the evaluation of reconstruction quality, base cameras are used to render images from 3D scenes, and corresponding PANO set serves as the reference. We employ three reference image quality evaluation metrics: structure similarity index measure (SSIM) , peak signal-to-noise ratio (PSNR), and learned perceptual image patch similarity (LPIPS)\coloredcite{zhang2018unreasonable}, to evaluate the quality of the rendered images. For the evaluation of rendering robustness, we render images using supplementary cameras, and employ traditional no-reference image quality assessment metrics: Natural Image Quality Evaluator (NIQE)\coloredcite{mittal2012making} and Blind/Referenceless Image Spatial Quality Evaluator (BRISQUE)\coloredcite{mittal2012no}.

Notably, taking into account the final file size of 3D-GS, the point cloud derived from our process is of a reduced size relative to the original RGB panoramic images. which leads to a loss of pixel information in the point clouds. Text2Room and LucidDreamer perform reconstruction based solely on the point clouds and consequently the reconstruction quality suffers loss. Although Text2NeRF could use RGB images from PANO set that is decoupled from depth information for supervision, the poor training efficiency of NeRF leads to excessively long reconstruction times. Our method, however, is capable of rapidly reconstructing from a single panorama and achieving excellent reconstruction quality. Additionally, All baseline methods do not adequately account for the robustness of rendering, which results in the presence of artifacts or missing regions within the reconstructed scenes, as shown in Fig.~\ref{fig:recon_compare}. Our method significantly enhances the quality of rendered images under supplementary view, effectively improving the rendering robustness of the reconstructed scene.

\begin{table}
        \caption{Quantitative Ablation Studies of Panorama Reconstruction from Single Panorama. We Evaluate the Effects of Important Components on Rendering Robustness. (\textbf{Best})\label{tab:recon_ablation}}
        \renewcommand{\arraystretch}{1.2}
	\centering
	\begin{tabular}{ccccc}
		\hline
		Method & NIQE\textdownarrow & BRISQUE\textdownarrow & TReS\textuparrow & MANIQA\textuparrow \\
		\hline
            \bf{Full} & 5.372 & \bf{32.372} & \bf{77.009} & \bf{0.402} \\
            w/o initial & \bf{5.369} & 32.598 & 76.491 & 0.399 \\
            w/o filter & 5.441 & 32.693 & 76.990 & 0.401 \\
            w/o initial \& filter & 5.419 & 32.625 & 76.367 & 0.398 \\
            \hline
	\end{tabular}
\end{table}

\noindent {\bf{Ablation Study.}}
Furthermore, we conduct ablation studies to ascertain the critical role of depth filtering of the point cloud and initial optimization using PCD set in the Pre Optimization stage in rendering robustness, as shown in Tab.~\ref{tab:recon_ablation}. To improve the discrimination, we incorporate two additional deep learning-based no-reference image quality assessment metrics: Manifold based Image Quality Assessment (ManIQA)\coloredcite{yang2022maniqa} and Training-free Referenceless Image Quality Evaluator (TReS)\coloredcite{golestaneh2022no}. The incorporation of these metrics allows for a more nuanced evaluation of image quality without the need for reference images. As Fig.~\ref{fig:recon_ablation} depicts, depth filtering of the point cloud and multi-view constraints of PCD set collectively contribute to reducing artifacts and inpainting missing regions, demonstrating the indispensable nature of each component.

\section{Limitations and Future Work}
Although our research has yielded commendable outcomes, several challenges persist with our current model. Panoramic image data are significantly scarcer compared to perspective image data, which poses a substantial challenge for the development of panorama generation methods. The data scarcity limits the capacity for effectively processing more complex text descriptions during generation. Combining of multiple diffusion models can expand the domain for generated panoramas, but can also result in compounded errors and an increase in stochastic variability. It is conceivable that forthcoming video generation models could contribute to producing more extensive and diverse panorama datasets.

Additionally, to ensure the efficiency of the 3D reconstruction, we have limited our approach to a two-stage reconstruction, with additional cameras introduced in the second stage to fill in missing areas. To further enhance the integrity of the scene and the robustness of the rendering, future work could increase the number of iterative inpainting stages. In addition, optimizing the camera setup strategy for each stage and refining the parameters of the optimization will be necessary to balance reconstruction quality and efficiency.

\section{Conclusion}
In this paper, we introduce HoloDreamer for generating highly consistent, fully enclosed 3D scenes with enhanced rendering robustness based on text descriptions. The pipeline consists of two proposed modules: Stylized Equirectangular Panorama Generation and Enhanced Two-Stage Panorama Reconstruction.
Our method not only refines the visual consistency and visual harmony but also enhances the integrity of the scenes and robustness of the rendering. The results of extensive experiments indicate that HoloDreamer represents a significant advancement in the field of 3D scene creation, offering a comprehensive solution that transforms text descriptions into intricate, immersive, and visually coherent 3D scenes.

{\appendix[Implement Details]
We implement HoloDreamer with the PyTorch framework. In the stage of panorama generation, we use the base model pre-trained in Diffusion360 \coloredcite{feng2023diffusion360}, which was fine-tuned using the dreambooth \coloredcite{ruiz2023dreambooth} training method in the SUN360 dataset \coloredcite{xiao2012recognizing} to produce the base panorama with a resolution of 1024x512. For style transfer on the panorama, we employ version 1.1 of ControlNet Lineart \coloredcite{zhang2023adding}, which is based on version 1.5 of Stable Diffusion \coloredcite{rombach2021highresolution}, to generate the stylized panorama with a resolution of 1536x728.
Subsequently, we refine the panorama by using the ControlNet Tile and Real-ESRGAN \coloredcite{wang2021realesrgan} following the refinement process in Diffusion360, achieving a detailed panorama of 6144x3072 resolution.


For depth estimation of the panorama, we balance quality and speed by initially downsampling the panorama to 4096x2048 resolution. We adhere to the strategies and parameter settings of the image projection and alignment stage in 360monodepth \coloredcite{rey2022360monodepth}. Disparity estimation is conducted using Depth Anything \coloredcite{depthanything}, a zero shot monocular relative depth estimation model, and then blended with frustum weights. Subsequently, ZoeDepth-NK \coloredcite{bhat2023zoedepth} is utilized to estimate metric depth to provide a reference for converting disparity map into a depth map.
The size of the raw point cloud $\boldsymbol{P_0}$ is the same as the resolution of the depth map, that is, 4096x2048, and the downsampled point cloud $\boldsymbol{P_s}$ is 1024x512. To obtain depth filtered point cloud $\boldsymbol{P_f}$, the point cloud is first downsampled to 2048x1024, and the threshold of depth gradient is set to 0.4. During the phase of the 3D Gaussian optimization, the camera's intrinsic parameters $\boldsymbol{K}$ are identical to the settings in LucidDreamer \coloredcite{chung2023luciddreamer}. All perspective images for supervision have the same resolution of 512x512. For the trajectory of the base cameras, we arrange a total of $38$ base views that provide a comprehensive coverage of a sphere, with each base camera corresponding to $4$ supplementary cameras positioned above, below, left and right. We use LaMa \coloredcite{suvorov2021resolution} to fill in the missing pixels of rendered images. The learning rate for the optimization of 3D Gaussian Splatting (3D-GS) is consistent with the original paper's \coloredcite{kerbl20233d} settings. The split and clone technique triggers every 100 iterations. The Pre Optimization stage involves an initial 2000 iterations for the optimization using PCD set and subsequently 2000 iterations for the optimization using PANO set. Furthermore, the Transfer Optimization stage consists of a total of 5,000 iterations. 

\begin{table}
        \setlength{\tabcolsep}{2pt}
        
        \caption{Quantitative Comparison of 360monodepth using MIDAS and Depth Anything on Replica360-4K at 4096×2048 with Frustum Blending and Multi-scale Deformable Alignment.(\textbf{Best})\label{tab:depth_com}}
        \renewcommand{\arraystretch}{1.2}
	\centering
	\begin{tabular}{cccccccc}
		\hline
		Method & \tiny AbsRel\textdownarrow & \tiny MAE\textdownarrow & \tiny RMSE\textdownarrow & \tiny RMSE-log\textdownarrow & \tiny $\delta\!\!<\!\!1.25$\textuparrow & \tiny $\delta\!\!<\!\!1.25\raisebox{0pt}{\textsuperscript{\fontsize{4pt}{12pt}\selectfont 2}}$\textuparrow & \tiny $\delta\!\!<\!\!1.25\raisebox{0pt}{\textsuperscript{\fontsize{4pt}{12pt}\selectfont 3}}$\textuparrow\\
		\hline
            MiDaS v2\coloredcite{Ranftl2022} & 0.153 & 0.346 & 0.579 & 0.082 & 0.810 & 0.949 & 0.982\\
            MiDaS v3\coloredcite{Ranftl2021} & 0.148 & 0.341 &  0.570 & 0.079 & 0.814 & 0.969 & 0.991\\
            \scriptsize Depth Anything\coloredcite{depthanything} & \textbf{0.116} & \textbf{0.281} & \textbf{0.481} & \textbf{0.063} & \textbf{0.894} & \textbf{0.984} & \textbf{0.995}\\
            \hline
	\end{tabular}
\end{table}

In addition, we compare the performance of 360monodepth using Depth Anything and MiDaS (in the original paper \coloredcite{rey2022360monodepth}) in Tab.\ref{tab:depth_com}, which proves that Depth Anything has better accuracy in panorama disparity estimation.}

\bibliographystyle{IEEEtran}
\bibliography{IEEEabrv,reference}

\begin{IEEEbiography}[{\includegraphics[width=1in,height=1.25in,clip,keepaspectratio]{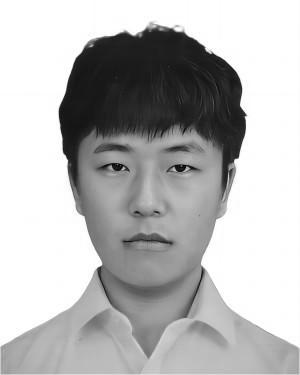}}]{Haiyang Zhou} is an undergraduate student at the School of Computer Science and Technology, Harbin Institute of Technology, Shenzhen, China.
He is currently interning at School of Electron and Computer Engineering, Peking University. His research interests include AIGC and 3D computer vision.
\end{IEEEbiography}

\begin{IEEEbiography}[{\includegraphics[width=1in,height=1.25in,clip,keepaspectratio]{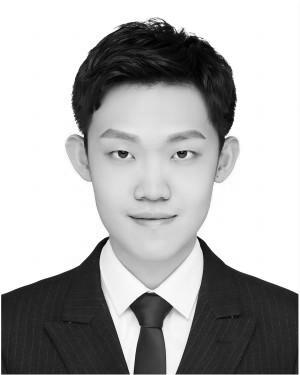}}]{Xinhua Cheng} recieved B.E. degree of computer science at College of Computer Science, Sichuan University. He is currently a PhD student of computer applications technology at School of Electron and Computer Engineering, Peking University, China. His research interests include 3D vision and AIGC, especially text-to-3D content creation and editing.
\end{IEEEbiography}

\begin{IEEEbiography}[{\includegraphics[width=1in,height=1.25in,clip,keepaspectratio]{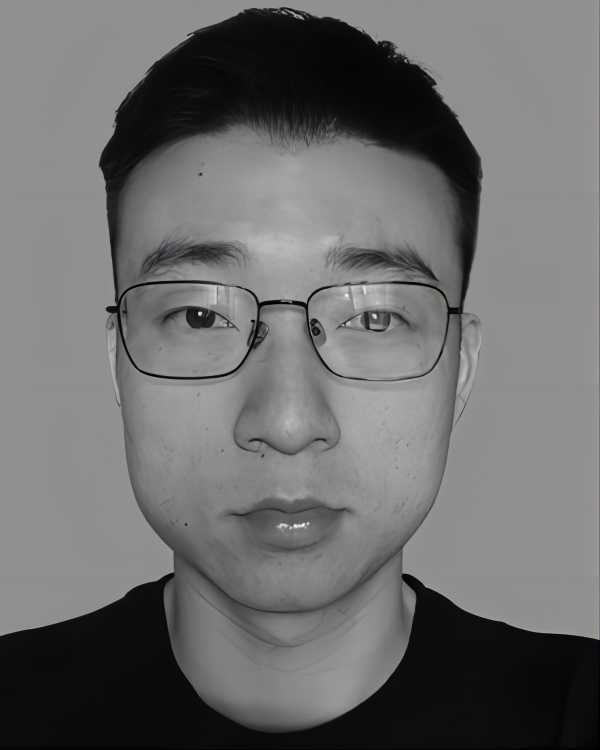}}]{Wangbo Yu}
received the B.E. degree in telecommunications engineering from Xidian University, Xi’an, China, in 2021. He is currently a PhD student with Peking University, Beijing, China. His research interests include computer vision, machine learning, and computer graphics.
\end{IEEEbiography}

\begin{IEEEbiography}[{\includegraphics[width=1in,height=1.25in,clip,keepaspectratio]{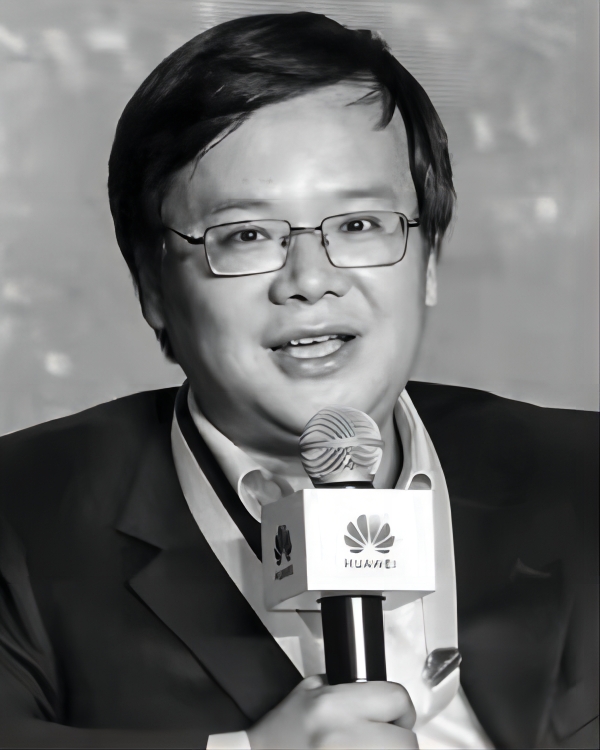}}]{Yonghong Tian}
(Fellow, IEEE) is currently the Dean of the School of Electronics and Computer Engineering, a Boya Distinguished Professor with the School of Computer Science, Peking University, China, and the Deputy Director of the Artificial Intelligence Research, Peng Cheng Laboratory, Shenzhen, China. He is the author or coauthor of over 350 technical papers in refereed journals and conferences. His research interests include neuromorphic vision, distributed machine learning, and AI for science. He is a TPC Member of more than ten conferences, such as CVPR, ICCV, ACM KDD, AAAI, ACM MM, and ECCV. He is a Senior Member of CIE and CCF and a member of ACM. He was a recipient of the Chinese National Science Foundation for Distinguished Young Scholars in 2018, two National Science and Technology Awards, and three ministerial-level awards in China. He received the 2015 Best Paper Award for \textit{EURASIP Journal on Image and Video Processing}, the Best Paper Award from IEEE BigMM 2018, and the 2022 IEEE SA Standards Medallion and SA Emerging Technology Award. He served as the TPC Co-Chair for BigMM 2015, the Technical Program Co-Chair for IEEE ICME 2015, IEEE ISM 2015, and IEEE MIPR 2018/2019, and the General Co-Chair for IEEE MIPR 2020 and ICME 2021. He was/is an Associate Editor of \textsc{IEEE Transactions on Circuits and Systems for Video Technology} from January 2018 to December 2021, \textsc{IEEE Transactions on Multimedia} from August 2014 to August 2018, \textit{IEEE Multimedia Magazine} from January 2018 to August 2022, and \textsc{IEEE Access} from January 2017 to December 2021. He co-initiated the IEEE International Conference on Multimedia Big Data (BigMM).
\end{IEEEbiography}

\begin{IEEEbiography}[{\includegraphics[width=1in,height=1.25in,clip,keepaspectratio]{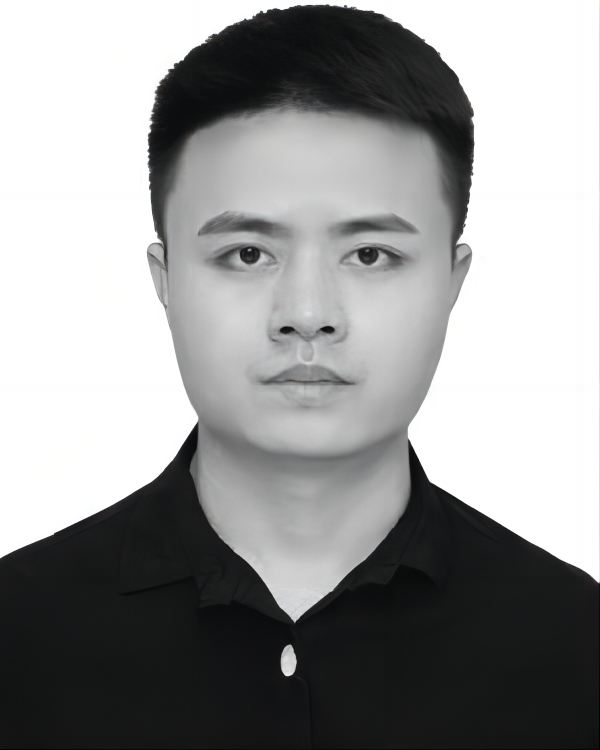}}]{Li Yuan}
received the B.E. degree from University of Science and Technology of China, in 2017, and the PhD degree from National University of Singapore, in 2021. He is currently a tenure-track assistant professor with School of Electrical and Computer Engineering with Peking University. He has published more than 40 papers on top conferences/journals. His research interests include deep learning, image processing, and computer vision.
\end{IEEEbiography}

\end{document}